\documentclass[runningheads]{llncs}
\usepackage{graphicx}
\usepackage{amsmath}
\usepackage{algorithm}
\usepackage{algorithmic}
\usepackage{booktabs}
\begin{document}
\title{Boosted Ensemble Learning based on Randomized NNs for Time Series Forecasting}

\titlerunning{Boosted Ensemble Learning based on RandNNs for TS Forecasting}
%
\author{Grzegorz Dudek\orcidID{0000-0002-2285-0327}}
\authorrunning{G. Dudek}
%
\institute{Electrical Engineering Faculty, \\ Czestochowa University of Technology, Czestochowa, Poland\\
\email grzegorz.dudek@pcz.pl}
\maketitle              
\begin{abstract}
Time series forecasting is a challenging problem particularly when a time series expresses multiple seasonality, nonlinear trend and varying variance. In this work, to forecast complex time series, we propose ensemble learning which is based on randomized neural networks, and boosted in three ways. These comprise ensemble learning based on residuals, corrected targets and opposed response. The latter two methods are employed to ensure similar forecasting tasks are solved by all ensemble members, which justifies the use of exactly the same base models at all stages of ensembling. Unification of the tasks for all members simplifies ensemble learning and leads to increased forecasting accuracy. This was confirmed in an experimental study involving forecasting time series with triple seasonality, in which we compare our three variants of ensemble boosting. The strong points of the proposed ensembles based on RandNNs are extremely rapid training and pattern-based time series representation, which extracts relevant information from time series. 

\end{abstract}
\section{Introduction}

Ensemble methods are considered to be a cornerstone of modern machine learning \cite{Ree18}. They are commonly used for regression and classification problems. 
Ensembling is also a very effective way of increasing the predictive power of forecasting models. Combining many base models improves the final forecasting accuracy as well as the stability of the response when compared to a single model approach.
Success in ensemble learning depends on the proper flexibility of the ensemble members and the trade-off between their performance and diversity \cite{Bro05}. It is also determined by the way learners are generated at the successive stages of ensembling and the method employed to combine them. 

The effectiveness of ensembling in forecasting is evidenced by the fact that in the most renowned forecasting competition, M4 \cite{Mar18}, of the 17 most accurate models, 12 used ensembling in some form \cite{Ati20}. The winning submission, which is a hybrid model combining exponential smoothing and long short-term memory, used three types of ensembling simultaneously \cite{Smyl20}: combining results of the stochastic training process, bagging, and combining multiple runs.

To improve the performance of ensemble learning many approaches have been proposed such as stacking \cite{Wol92}, bagging \cite{Bre96}, boosting \cite{Dru99}, negative correlation learning \cite{Chen09}, snapshot ensembles \cite{Hua17},  
and horizontal and vertical ensembles developed for deep learning \cite{Xie13}. Boosting, which this work focuses on, is a general ensemble technique that involves sequentially adding base models to the ensemble where subsequent models correct the performance of prior models. This approach is very effective
as evidenced by the high ranking positions of boosted models such as XGBoost \cite{Chen16}, i.e. ensemble of decision trees with regularized gradient boosting, in competitions such as those organized by Kaggle. Boosting also has many applications in the forecasting field. Some examples are: \cite{Zie16}, where an ensemble of boosted trees is used for bankruptcy prediction; 
\cite{Ni20}, where XGBoost is combined with a Gaussian mixture model for monthly streamflow forecasting;
\cite{Li19}, where AdaBoost is applied as a component of a hybrid model for multi-step wind speed forecasting;
and \cite{Mit22}, where a natural gradient boosting algorithm is applied for solar power probabilistic forecasting. 
This last work highlights a valuable advantage of ensembling. It can produce probabilistic forecasts, i.e. the distribution of the forecasted variable in the future.

In this study, we propose a boosted version of our ensemble of randomized neural networks (RandNNs) for complex time series forecasting \cite{Dud21}. In \cite{Dud21}, we focused on strategies for controlling the diversity of ensemble members. The members were trained independently. Here, we construct an ensemble sequentially. When a new member is added to the ensemble, it learns by taking into account the results of the ensemble members so far. We consider three methods of boosting. The contribution of this study is threefold: 


\begin{enumerate}
    \item 
    We propose new methods of ensemble boosting: ensemble learning based on corrected targets and ensemble learning based on opposed response. They are both employed to ensure similar tasks for all ensemble members. This unification of the tasks justifies the use of identical base models in terms of architecture and hyperparameters at all stages of ensembling.

    \item
    We develop three ensemble learning approaches for forecasting complex time series with multiple seasonality. They are based on RandNNs, pattern-based time series representation and three methods of boosting: based on residuals, corrected targets and opposed response.
 
    \item 
    We empirically compare the performance of the proposed boosted ensemble methods on challenging short-term load forecasting problems with triple seasonality, and conclude that the opposed response-based approach outperforms its competitors in terms of forecasting accuracy and sensitivity to hyperparameters.      
    
\end{enumerate} 

The rest of the work is organized as follows. In Section II, we present a base model, RandNN. Details of the proposed three methods of boosted ensemble learning are described in Section III. The experimental framework used to evaluate and compare the proposed ensemble methods is described in Section IV. Finally, Section V concludes the work.

\section{Base Forecasting Model - RandNN}

As a base model, we use a single hidden layer feedforward NN with $m$ logistic sigmoid hidden nodes \cite{Dud21a}. In randomized learning, the weights of hidden nodes are selected randomly from a uniform distribution and symmetrical interval $U=[-u,u]$. The biases of these nodes are calculated, according to recent research \cite{Dud19}, based on the weights as follows: $b_j = -\mathbf{a}_j^T\mathbf{x}^*_j$, where $\mathbf{a}_j$ is the vector of weights for the $j$-th hidden node, and $\mathbf{x}^*_j$ is one of the training patterns selected at random (see \cite{Dud19} and \cite{Dud20} for details, justification and other variants). This way of generating hidden nodes places the sigmoids into the input feature space limited to some hypercube, avoiding their saturated parts. Moreover, the sigmoids are distributed according to the data distribution. These improve the aproximation and generalization abilities of the model \cite{Dud19}, \cite{Dud20}. 

The hidden node sigmoids are combined linearly by the output nodes: $\varphi_k(\mathbf{x})=\sum_{j=1}^{m}\beta_{j,k}h_j(\mathbf{x})$, where $h_j(\mathbf{x})$ is the output of the $j$-th hidden node, and $\beta_{j,k}$ is the weight between $j$-th hidden and $k$-th output nodes. The only learnable parameters are the output weights. They are calculated from $\boldsymbol{\beta} = \mathbf{H}^+\mathbf{Y}$, where $\mathbf{H}$ is a matrix of the hidden layer outputs, $\mathbf{H}^+$ denotes its Moore–Penrose generalized inversion, and $\mathbf{Y}$ is a~matrix of target output patterns. $\mathbf{H}$ is a nonlinear feature mapping from $n$-dimensional input space to $m$-dimensional projection space (usually $m \gg n$). Note that this projection is random. Due to the fixed parameters of the hidden nodes, the optimization problem in randomized learning (selection of weights $\boldsymbol{\beta}$) becomes convex and can be easily solved by the standard least-squares method. 

The forecasting model based on RandNN, which we adapt from \cite{Dud21a}, has two more components: encoder and decoder. 
Let us consider time series $\{E_k\}_{k=1}^K$ with multiple seasonality. The encoder transforms this series into input and output patterns expressing seasonal sequences of the shortest length. The input patterns, $\mathbf{x}_i = [x_{i,1}, …, x_{i,n}]^T$, represent sequences $ \mathbf{e}_i = [E_{i,1},…, E_{i,n}]^T $, while the output patterns, $\mathbf{y}_i = [y_{i,1},…, y_{i,n}]^T$, represent forecasted sequences $ \mathbf{e}_{i+\tau} = [E_{i+\tau,1},…, E_{i+\tau,n}]^T $, where $n$ is a period of the seasonal cycle (e.g. 24 hours for daily seasonality), $i=1, ..., K/n$ is the sequence number, and $\tau \geq 1$ is a forecast horizon. The patterns are defined as follows:   

\begin{equation}\label{eq1}
\mathbf{x}_i = \frac{\mathbf{e}_i-\overline{e}_i}{\widetilde{e}_i},
\qquad
\mathbf{y}_i = \frac{\mathbf{e}_{i+\tau}-\overline{{e}}_i}{\widetilde{{e}}_i}
\end{equation}
where $\overline{{e}}_i$ is the mean value of sequence $\mathbf{e}_i$, and $\widetilde{e}_i = \sqrt{\sum_{t=1}^{n} (E_{i,t}-\overline{e}_i)^2}$ is a measure of sequence $\mathbf{e}_i$ dispersion.

Input patterns $\mathbf{x}_i$ represent successive seasonal sequences which are centered and normalized. They have a zero mean, and the same variance and unity length. Thus they are unified and differ only in shape. In contrast, patterns $\mathbf{y}_i$ are not globally unified and can express additional seasonality, e.g. when time series include both daily and weekly seasonalities, the former is expressed in the y-pattern shape, while the latter is expressed in y-pattern level and dispersion (compare x- and y-patterns in Fig. 2 and see discussion in \cite{Dud21a}). In the case of such seasonalities, we build forecasting models that learn from data representing the same days of the week, e.g. for test query pattern $\mathbf{x}$ representing Monday, training set $\Phi$ consists of x-patterns representing all historical Mondays from the data and y-pattern representing the Tuesdays following them  (assuming $\tau = 1$). 

The decoder based on the y-pattern forecasted by the network, $\widehat{\mathbf{y}}$, and coding variables describing the query sequence, $\widetilde{{e}}$ and $\overline{{e}}$, using transformed equation \eqref{eq1} for $\mathbf{y}$, calculates the forecasted seasonal sequence:

\begin{equation}\label{eq3}
\widehat{\mathbf{e}} = \widehat{\mathbf{y}}\widetilde{{e}}+\overline{{e}} 
\end{equation}

Remarks:

\begin{enumerate}
    \item 
RandNN was designed for forecasting time series with multiple seasonalities. In the case of one seasonality, y-patterns express only one seasonality, and there is no need to decompose the forecasting problem. In the case of no seasonality, the input pattern length should be selected experimentally, while the y-pattern length is equal to the forecast horizon.

    \item
The bounds of the interval for random weights, $u$, correspond to the maximum sigmoid slope. To increase interpretability, let us express the bounds $u$ using the slope angles $\alpha_{max}$ \cite{Dud20}: $u=4 \tan \alpha_{max}$, and treat $\alpha_{max}$ as a hyperparameter. The second hyperparameter is the number of hidden nodes, $m$. Both hyperparameters decide about the bias-variance tradeoff of the model and should be tuned to the complexity of the target function.
 
    \item 
Advantages of RandNN are extremly fast training and the simplification of the forecasting task due to pattern representation. 
In \cite{Dud21a}, it was shown that RandNN can compete with fully-trained NN in terms of forecasting accuracy, but is much faster to train.   

\end{enumerate} 

\section{Boosting of Ensemble Learning}

\subsection{Ensemble Learning Based on Residuals}




An ensemble based on residuals, which is a simplified variant of a gradient boosting algorithm \cite{Mas99}, is constructed sequentially. In the first step, a base model learns on the training set $\{(x_i,y_i)\}_{i=1}^N$ (we consider a scalar input and output for simplicity) and fits to original data. Let us denote this model $f_1(x)$ and the ensemble model including one member $F_1(x)=f_1(x)$. In the second step, the second base model is added, $f_2(x)$, such that the sum of this model with the previous one, $F_2(x)=F_1(x)+f_2(x)$, is the closest possible to the target $y$. In the successive steps, further base models are added with the same expectation. In the $k$-th step, the function fitted by the ensemble is $F_k(x)=F_{k-1}(x)+f_k(x)$. Thus, the error in this step, $MSE_k=\frac{1}{N}\sum_{i=1}^{N} \left(y_i-F_k(x_i)\right)^2$, can be written as:

\begin{equation}
MSE_k=\frac{1}{N}\sum_{i=1}^{N}\left(f_k(x_i)-r_{k-1,i}\right)^2
\label{MSE}
\end{equation}
where $r_{k-1,i} = y_i-F_{k-1}(x_i)$ is a residual between the target and the response of the ensemble of $k-1$ members. 

Equation \eqref{MSE} clearly shows that the base model added to the ensemble at the $k$-th stage fits to the residuals between the target and the ensemble built at stage $k-1$. So, it attempts to correct the errors of its predecessors. The training set for the base model in the $k$-th stage is $\{(x_i,r_{k-1,i})\}_{i=1}^N$. The final ensemble response is the sum of all member responses: $F_K(x)=\sum_{k=1}^{K} f_k(x)$.   

\begin{algorithm}[h]
	\caption{EnsR: Boosting based on residuals}
	\label{alg1}
	\begin{algorithmic}
		\STATE {\bfseries Input:} Base model $f$ (RandNN), Training set $\Phi = \{(\mathbf{x}_i, \mathbf{y}_i)\}_{i=1}^N$, Ensemble size $K$\\

		\STATE {\bfseries Output:} RandNN ensemble $F_K$\\    
		\STATE {\bfseries Procedure:}\\
		\FOR{$k=1$ {\bfseries to} $K$}
		\STATE Learn $f_k$ based on $\Phi$ \\
		\STATE Calculate ensemble response $F_k(\mathbf{x}_i)=\sum_{l=1}^{k} f_l(\mathbf{x}_i), i = 1, ..., N$\\
        \STATE Determine residuals $\mathbf{r}_{k,i}=\mathbf{y}_i-F_k(\mathbf{x}_i), i = 1, ..., N$\\ 
        \STATE Modify training set $\Phi = \{(\mathbf{x}_i, \mathbf{r}_{k,i})\}_{i=1}^N$ \\
		\ENDFOR
	\end{algorithmic}
\end{algorithm}

\begin{figure}[h]
	\centering
	\includegraphics[width=0.22\textwidth]{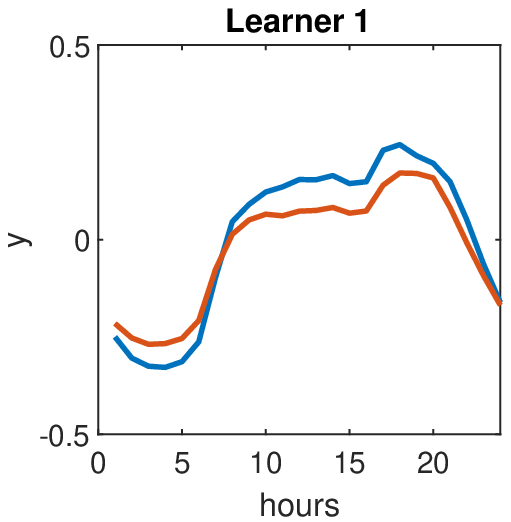}
    \includegraphics[width=0.22\textwidth]{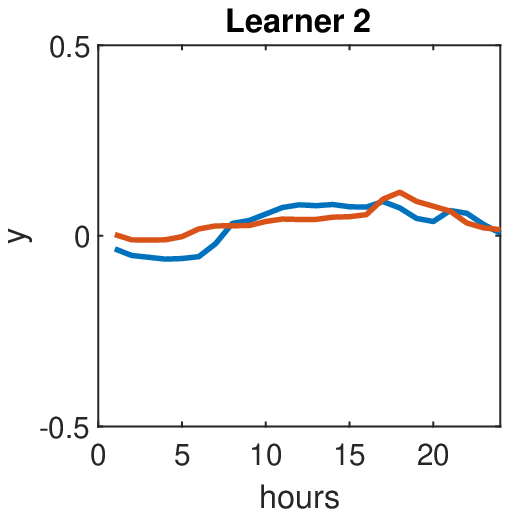}
    \includegraphics[width=0.25\textwidth]{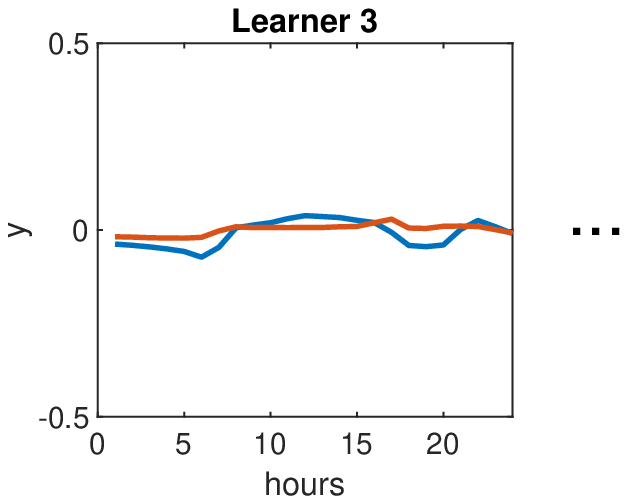}
    \includegraphics[width=0.22\textwidth]{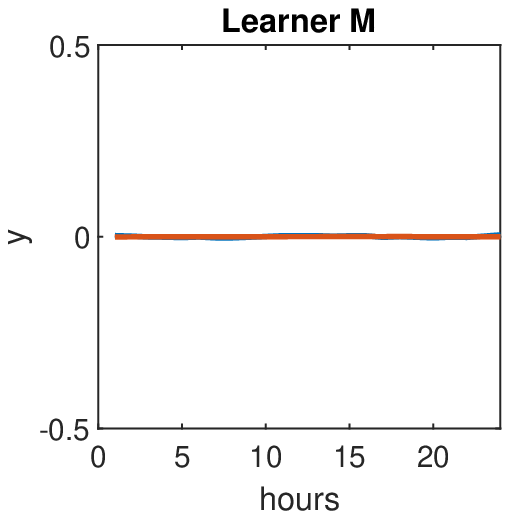}
    \includegraphics[width=0.45\textwidth]{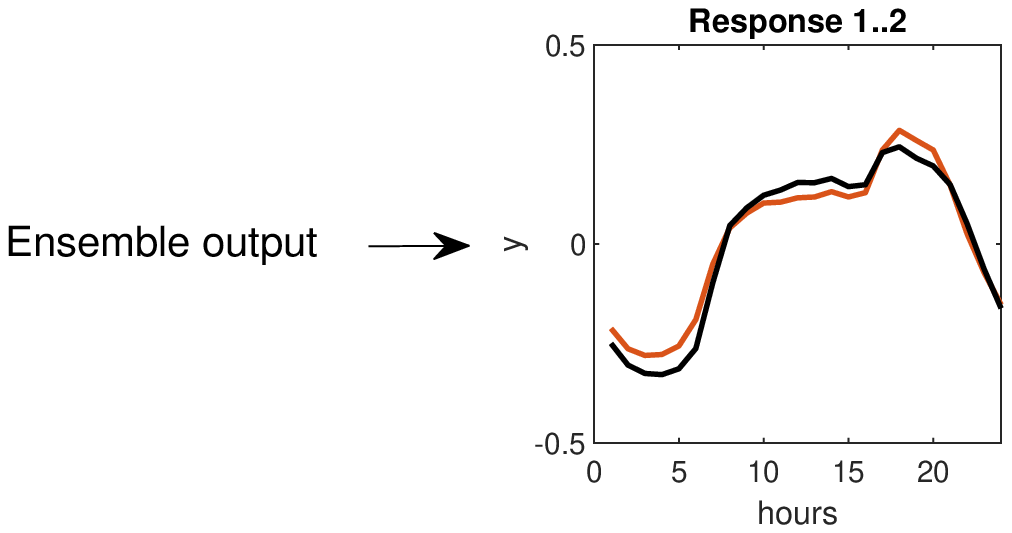}
    \includegraphics[width=0.25\textwidth]{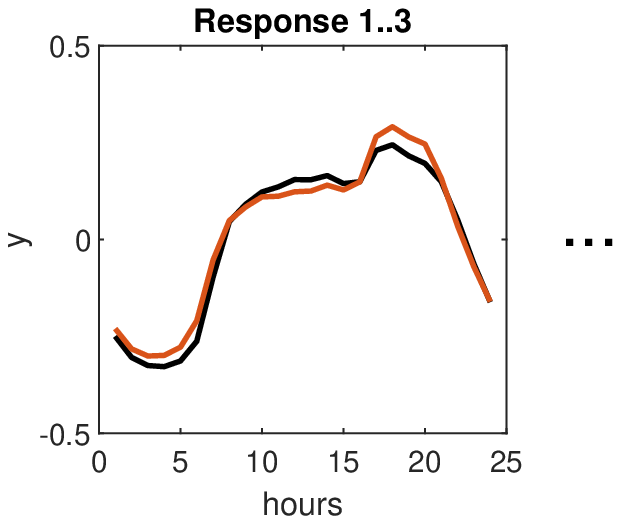}
    \includegraphics[width=0.22\textwidth]{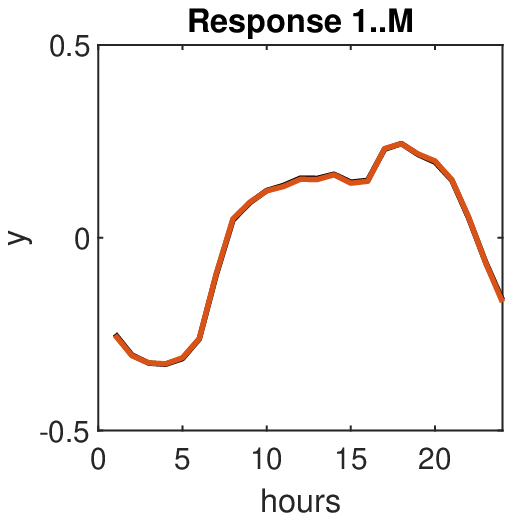}
    \caption{EnsR: Responses produced by individual learners (upper panel) and the ensemble (lower panel). Target responses of individual learners in blue, target responses of the ensemble in black, real responses in red.} 
    \label{FigEnsR}
\end{figure}

Algorithm \ref{alg1} and Fig. \ref{FigEnsR} demonstrate the process of building an ensemble based on residuals (EnsR) for pattern-based forecasting. Learner 1 (RandNN) learns the original target function on $\Phi$. Each subsequent learner learns the residuals between the target patterns $\textbf{y}$ and the aggregated outputs of its predecessors shown in the lower panel of Fig. \ref{FigEnsR}. Note that in EnsR, each learner can have a different problem to solve, expressing different features. Learner 1 learns the specific patterns of seasonal cycles $\textbf{y}$, while the next learners learn completely different tasks, i.e. the residuals, which do not have such distinct patterns as $\textbf{y}$ and have a large stochastic component. This inconsistency between the problems solved by the learners can affect negatively the final result, especially in the common case of using identical base models (the same architecture and hyperparameters) at every stage of ensembling.

\subsection{Ensemble Learning Based on Corrected Targets}

To unify the problems solved by the learners at successive stages of ensembling, we modify the EnsR framework as follows. During the sequential process, at stage $k$, the base model $f_k(x)$ is added to the ensemble. Ensemble response at this stage is the average of $k$ learners: $F_k(x)=\frac{1}{k}\sum_{l=1}^{k} f_l(x)$. The loss function can be expressed as:

\begin{equation}
\begin{split}
MSE_k & =\frac{1}{N}\sum_{i=1}^{N}\left(y- \left(\frac{1}{k}\sum_{l=1}^{k-1} f_l(x) + \frac{f_k(x)}{k}\right)\right)^2\\
& =\frac{1}{Nk}\sum_{i=1}^{N}\left(f_k(x)-\left(ky-\sum_{l=1}^{k-1} f_l(x)\right)\right)^2
\end{split}
\label{MSE2}
\end{equation}

As can be seen from this equation, the difference between $ky$ and the sum of the previous learners is the new target to which the $k$-th learner is fitted. This target expresses pattern $y$ corrected by aggregated residuals of the previous learners: $y+\sum_{l=1}^{k-1} (y-f_l(x))$. If the aggregated residuals are much smaller compared to the y-pattern (we expect this), the new targets at all stages of ensembling have a similar shape to the y-pattern. So, all learners have similar tasks to solve and it is justified for them to have the same architecture and hyperparameters. This unburdens us from the awkward and time-consuming task of selecting the optimal model at each stage of ensembling.

\begin{figure}[h]
	\centering
	\includegraphics[width=0.22\textwidth]{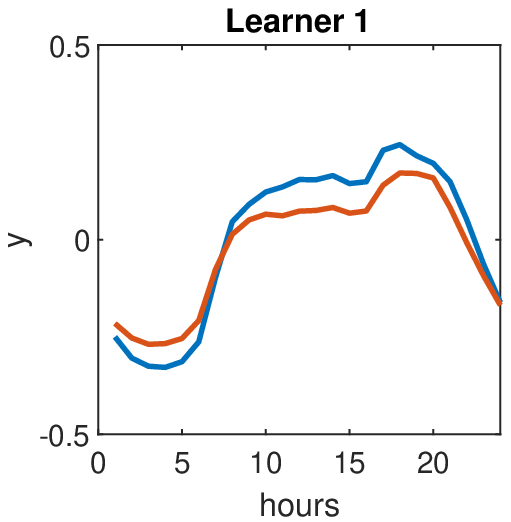}
    \includegraphics[width=0.22\textwidth]{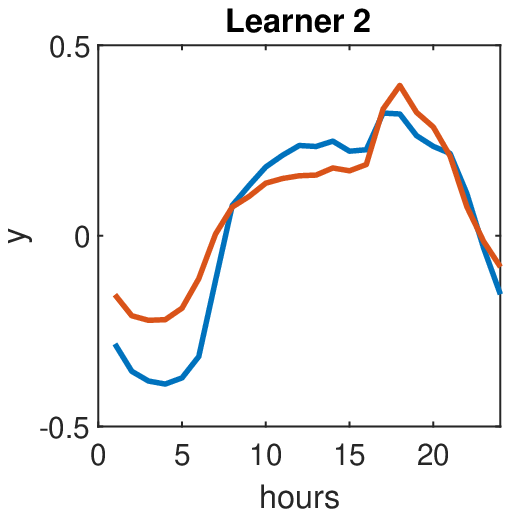}
    \includegraphics[width=0.25\textwidth]{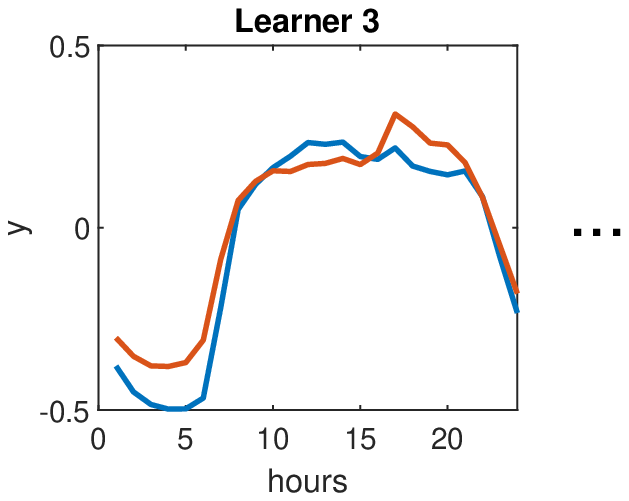}
    \includegraphics[width=0.22\textwidth]{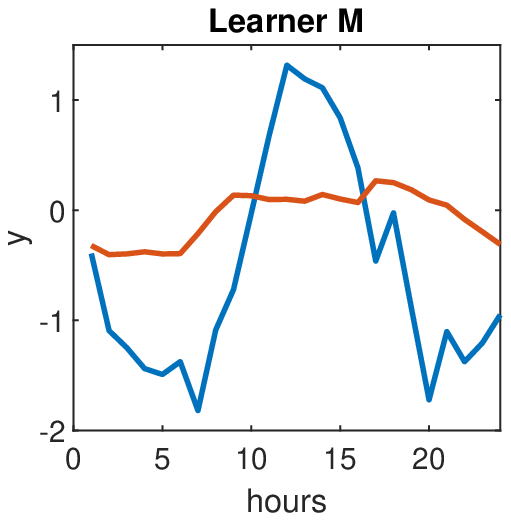}
    \includegraphics[width=0.45\textwidth]{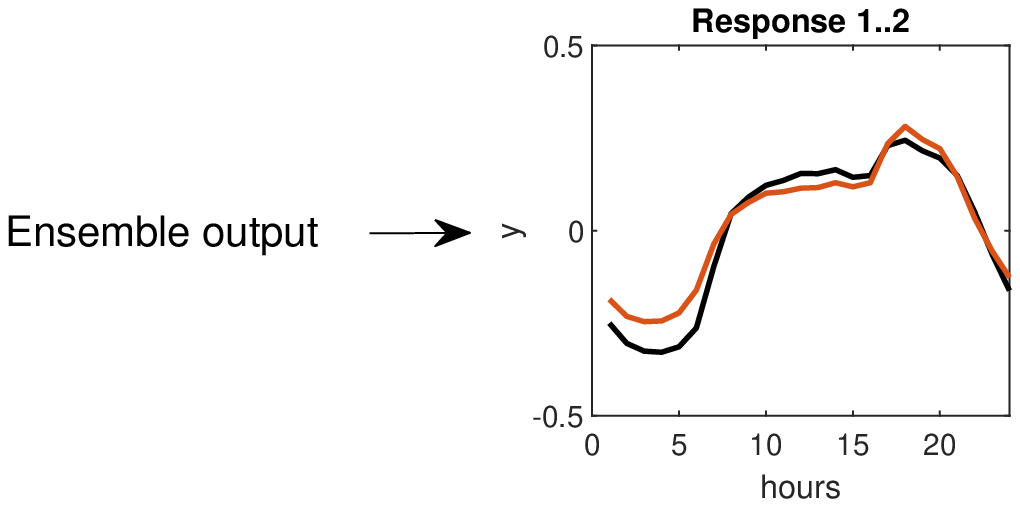}
    \includegraphics[width=0.25\textwidth]{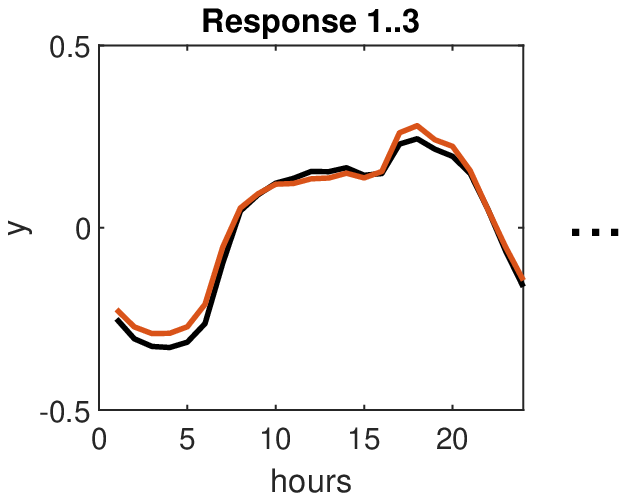}
    \includegraphics[width=0.22\textwidth]{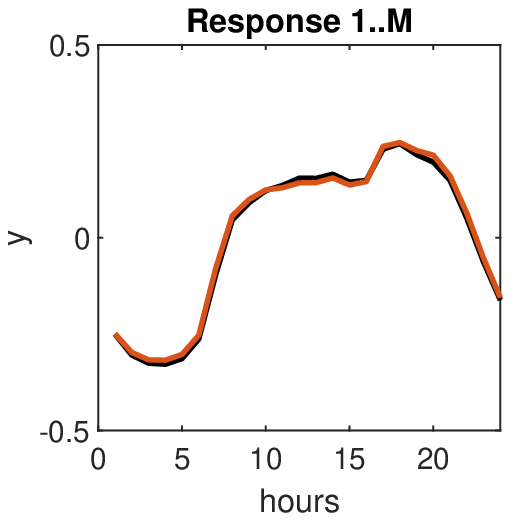}
    \caption{EnsCT: Responses produced by individual learners (upper panel) and the ensemble (lower panel). Target responses of individual learners in blue, target responses of the ensemble in black, real responses in red.} 
    \label{FigEnsR2}
\end{figure}

Algorithm \ref{alg2} summarizes ensemble learning based on corrected targets (EnsCT) for pattern-based forecasting. Fig. \ref{FigEnsR2} shows the targets for learners at successive stages, learners' outputs and the ensemble outputs. It was observed that at the initial stages of ensembling, the targets express the y-pattern shape, but this shape degrades gradually in the later stages - see the target for $K$-th learner in Fig. \ref{FigEnsR2}. Thus, the proposed EnsCT only partially solves the problem of inconsistency between tasks learned at the successive stages of ensembling.

\begin{algorithm}[h]
	\caption{EnsCT: Boosting based on corrected targets}
	\label{alg2}
	\begin{algorithmic}
		\STATE {\bfseries Input:} Base model $f$ (RandNN), Training set $\Phi = \{(\mathbf{x}_i, \mathbf{y}_i)\}_{i=1}^N$, Ensemble size $K$\\

		\STATE {\bfseries Output:} RandNN ensemble $F_K$\\    
		\STATE {\bfseries Procedure:}\\
		\FOR{$k=1$ {\bfseries to} $K$}
		\STATE Learn $f_k$ based on $\Phi$ \\
		\STATE Calculate ensemble response $F_k(\mathbf{x}_i)=\frac{1}{k}\sum_{l=1}^{k} f_l(\mathbf{x}_i), i = 1, ..., N$\\
        \STATE Determine corrected targets $\mathbf{y}'_{k,i}=(k+1)\mathbf{y}_i-kF_k(\mathbf{x}_i), i = 1, ..., N$\\ 
        \STATE Modify training set $\Phi = \{(\mathbf{x}_i, \mathbf{y}'_{k,i})\}_{i=1}^N$ \\
		\ENDFOR
	\end{algorithmic}
\end{algorithm}

\subsection{Ensemble Learning Based on Opposed Response}

To prevent the degradation of the target shapes in the subsequent steps of ensembling, we propose ensemble learning based on opposed response (EnsOR).
The first step of the ensemble building procedure is the same as in EnsR and EnsCT. 
At this stage, the base model $f_1(x)$ learns on original training set $\Phi$. The ensemble response is $F_1(x)=f_1(x)$. 
The residual is calculated, $r_1 = y-F_1(x)$, and the "opposed" response pattern is determined as follows:

\begin{equation}
\hat{y}'_1=y+r_1=2y-F_1(x)
\label{Opp}
\end{equation}

\begin{figure}
	\centering
	\includegraphics[width=0.5\textwidth]{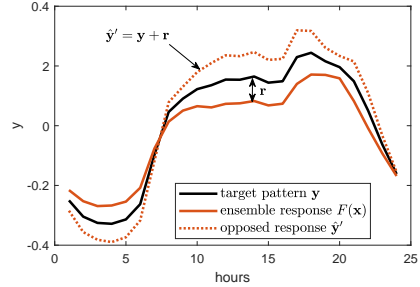}
    \caption{Construction of the opposed response pattern.}
    \label{figOP}
\end{figure}

The opposed response pattern expresses the target pattern augmented by the opposed error produced by the ensemble (see Fig. \ref{figOP}). The average of the response pattern $F_1(x)$ and the opposed response pattern $\hat{y}'_1$ gives the target pattern $y$. In the next step, base model $f_2(x)$ learns on training set $\{(x_i,\hat{y}'_{1,i})\}_{i=1}^N$. Thus it learns the opposed response to reduce the ensemble residual. The ensemble response is calculated as the average of learners: $F_2(x)=\frac{f_1(x)+f_2(x)}{2}$. These operations are repeated in the following steps (see Algorithm \ref{alg3}). Namely, in step $k$, the opposed response pattern determined at stage $k-1$, $\hat{y}'_{k-1}=2y-F_{k-1}(x)$, becomes the target pattern for learner $f_k(x)$. The ensemble response is the average of $k$ learners and the loss function at stage $k$ takes the form:

\begin{equation}
MSE_k=\frac{1}{N}\sum_{i=1}^{N}\left(f_k(x_i)-\hat{y}'_{k-1,i}\right)^2
\label{MSE2}
\end{equation}
where $\hat{y}'_{k-1,i}=y_i+r_{k-1,i}=2y_i-F_{k-1}(x_i)$ is the opposed response pattern at stage $k-1$.

\begin{algorithm}[h]
	\caption{EnsOR: Boosting based on opposed response}
	\label{alg3}
	\begin{algorithmic}
		\STATE {\bfseries Input:} Base model $f$ (RandNN), Training set $\Phi = \{(\mathbf{x}_i, \mathbf{y}_i)\}_{i=1}^N$, Ensemble size $K$\\

		\STATE {\bfseries Output:} RandNN ensemble $F_K$\\    
		\STATE {\bfseries Procedure:}\\
		\FOR{$k=1$ {\bfseries to} $K$}
		\STATE Learn $f_k$ based on $\Phi$ \\
		\STATE Calculate ensemble response $F_k(\mathbf{x}_i)=\frac{1}{k}\sum_{l=1}^{k} f_l(\mathbf{x}_i), i = 1, ..., N$\\
        \STATE Determine opposed response $\hat{\mathbf{y}}'_{k,i}=2\mathbf{y}_i-F_k(\mathbf{x}_i), i = 1, ..., N$\\ 
        \STATE Modify training set $\Phi = \{(\mathbf{x}_i, \hat{\mathbf{y}}'_{k,i})\}_{i=1}^N$ \\
		\ENDFOR
	\end{algorithmic}
\end{algorithm}

\begin{figure}[h]
	\centering
	\includegraphics[width=0.22\textwidth]{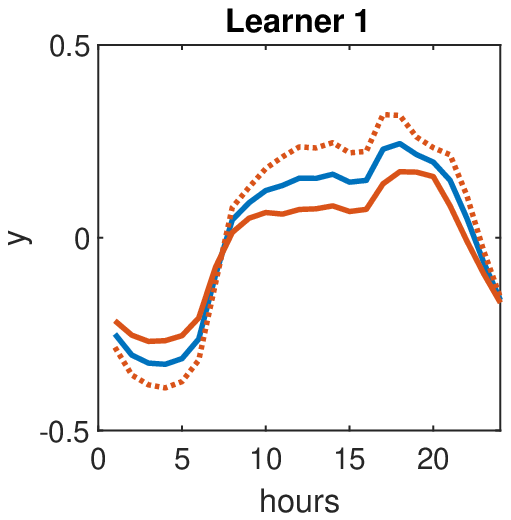}
    \includegraphics[width=0.22\textwidth]{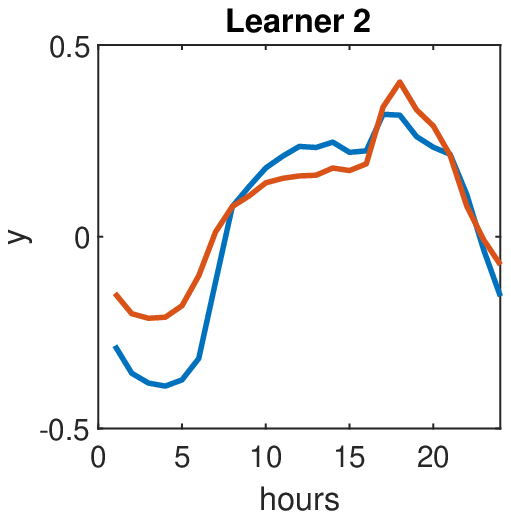}
    \includegraphics[width=0.25\textwidth]{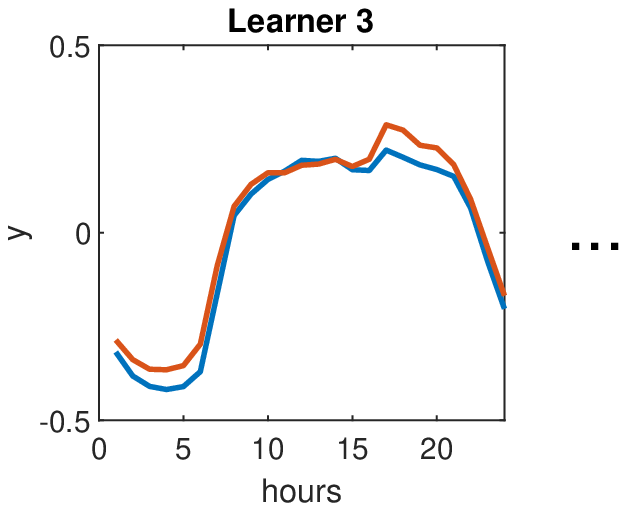}
    \includegraphics[width=0.22\textwidth]{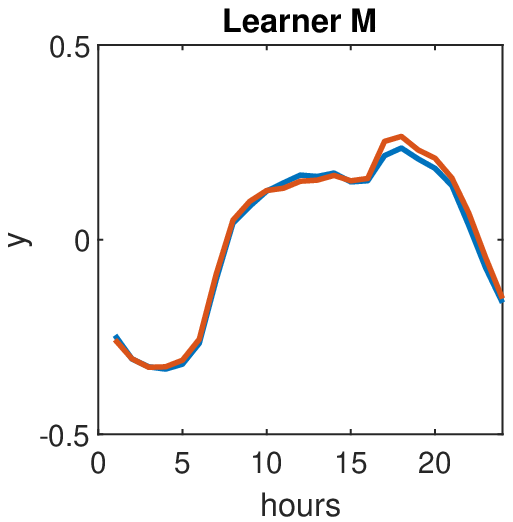}
    \includegraphics[width=0.45\textwidth]{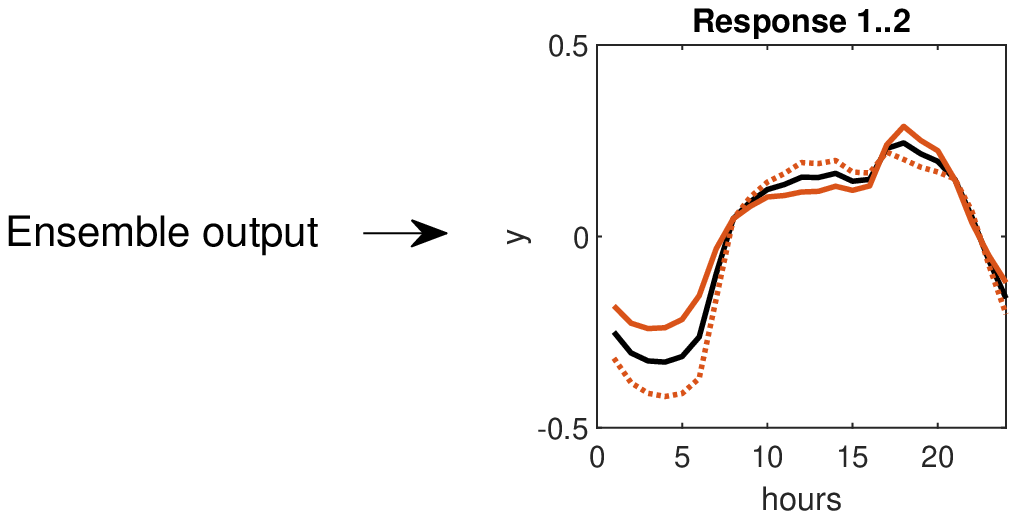}
    \includegraphics[width=0.25\textwidth]{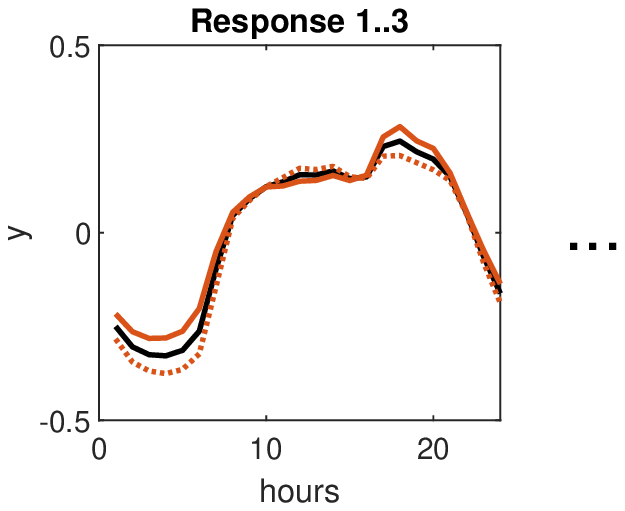}
    \includegraphics[width=0.22\textwidth]{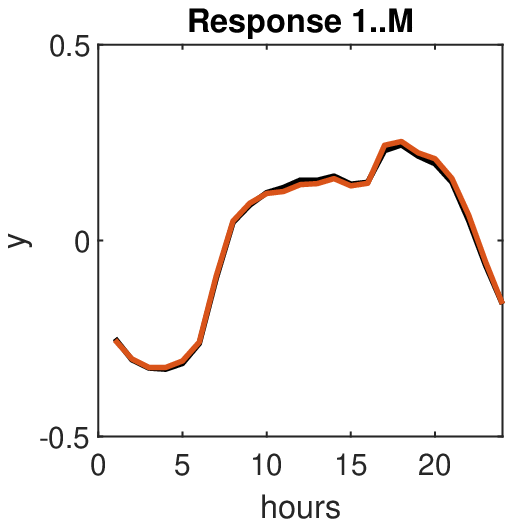}
    \caption{EnsOR: Responses produced by individual learners (upper panel) and the ensemble (lower panel). Target responses of individual learners in blue, target responses of the ensemble in black, real responses in red, and opposed responses in dashed red.} 
    \label{figEnsOR}
\end{figure}

Fig. \ref{figEnsOR} shows learners' and ensemble responses in the following steps of ensembling. Dashed lines express the opposed responses which become targets for the based models in the next steps (blue lines). Note that the shape of the initial target pattern $\mathbf{y}$ is maintained until the last stage. Thus, all learners learn on similar data, and using identical base models at each stage of ensembling means no objections can be raised, unlike in the case of EnsR.

\section{Experimental Study}

In this section, we verify our proposed methods of ensemble boosting on four time series forecasting problems. These comprise short-term electrical load forecasting problems for four European countries: Poland (PL), Great Britain (GB), France (FR) and Germany (DE) (data was collected from \url{www.entsoe.eu}). The hourly load time series express three seasonalities: yearly, weekly and daily. The data period is four years, from 2012 to 2015. Atypical days such as public holidays were excluded from the data (between 10 and 20 days a year). The forecasting problem is to predict the load profile (24 hourly values) for each day of 2015 based on historical data. The forecast horizon is one day, $\tau=1$. The number of ensemble members was $K=50$. As a performance metric we use mean absolute percentage error (MAPE).

\begin{figure}[h]
	\centering
	\includegraphics[width=0.7\textwidth]{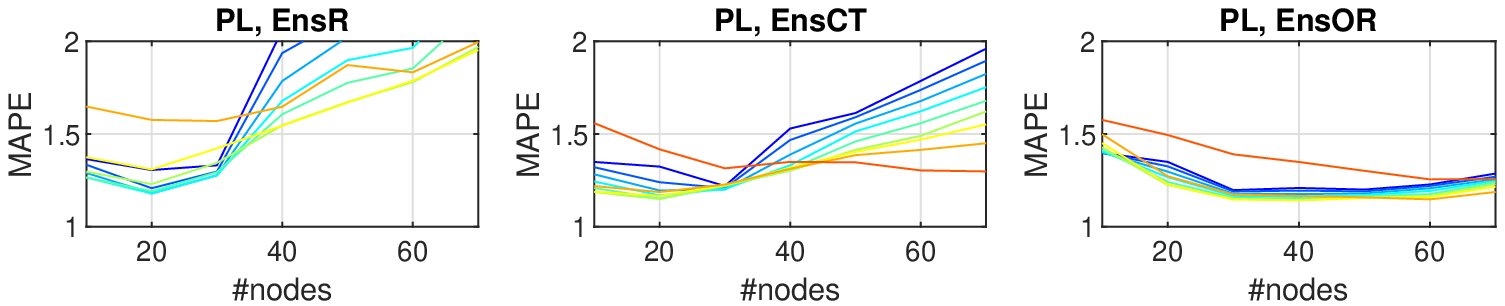}
	\includegraphics[width=0.042\textwidth]{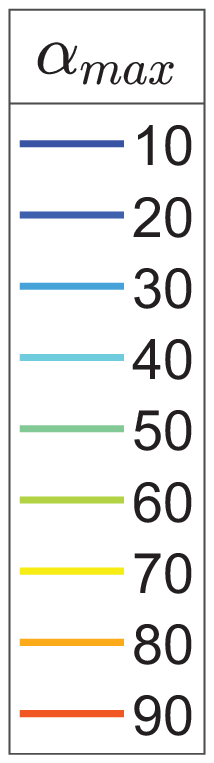}
    \includegraphics[width=0.7\textwidth]{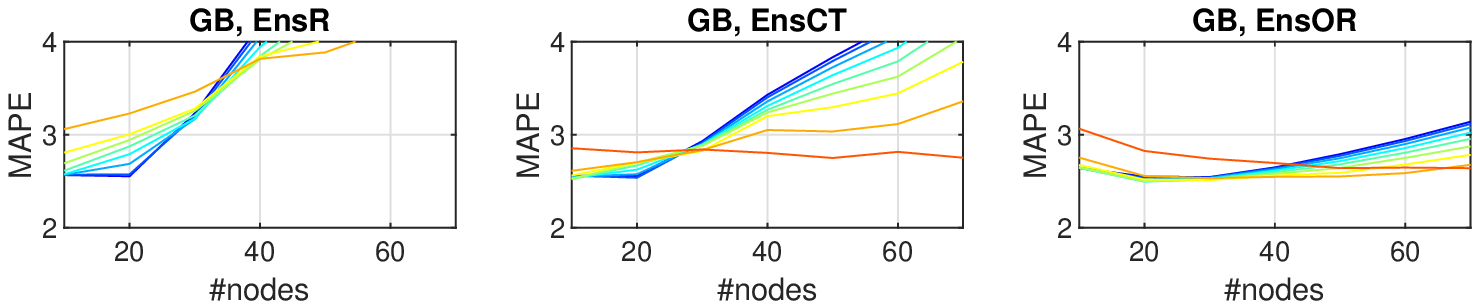}
    \includegraphics[width=0.042\textwidth]{empty.eps}
    \includegraphics[width=0.7\textwidth]{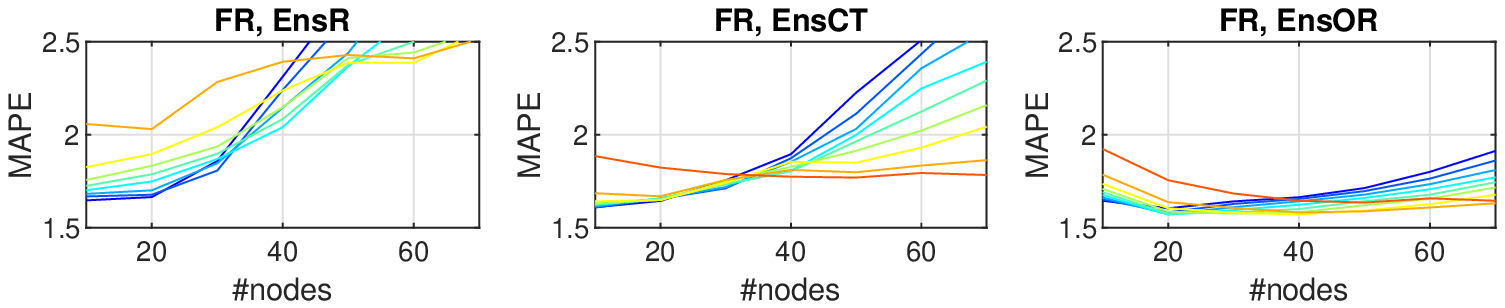}
    \includegraphics[width=0.042\textwidth]{empty.eps}
    \includegraphics[width=0.7\textwidth]{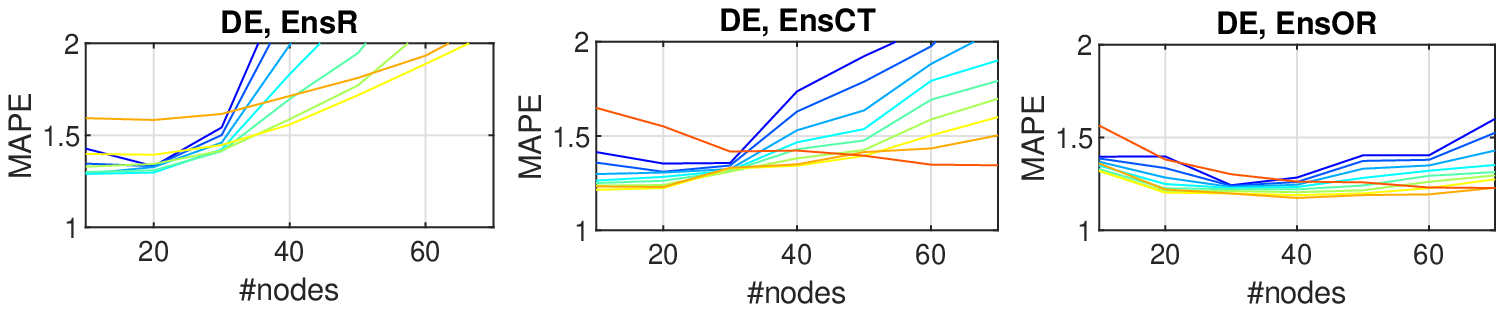}
    \includegraphics[width=0.042\textwidth]{empty.eps}
    \caption{Ensemble sensitivity to RandNN hyperparameters.} 
    \label{figSens}
\end{figure}

In the first experiment, we evaluate ensemble sensitivity to the base model hyperparameters: number of hidden nodes $m$ and size of the interval for hidden weights $\alpha_{max}$. Fig. \ref{figSens} shows the impact of hyperparameters on the test MAPE. Note that EnsR is the most sensitive to hyperparameters, while EnsOR is the least sensitive. 
To quantitatively compare the sensitivities of the ensemble variants, as a rough measure of sensitivity to a given hyperparameter, we define standard deviation of the test MAPE for the ensemble with optimal values of other hyperparameters:

\begin{equation}
S_m = Std(MAPE(\mathbf{y},F(\mathbf{x},\alpha^*_{max},m))) 
\label{Sm}
\end{equation}
\begin{equation}
S_{\alpha_{max}} = Std(MAPE(\mathbf{y},F(\mathbf{x},\alpha_{max},m^*)))
\label{Sa}
\end{equation}
where the optimal hyperparameter values are marked with asterisks.

\begin{figure}[h]
	\centering
	\includegraphics[width=0.24\textwidth]{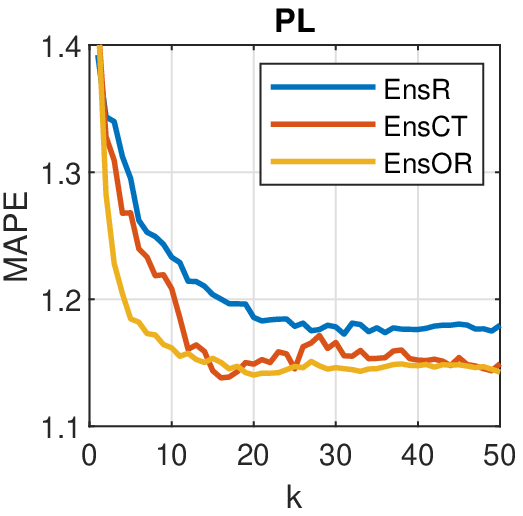}
    \includegraphics[width=0.24\textwidth]{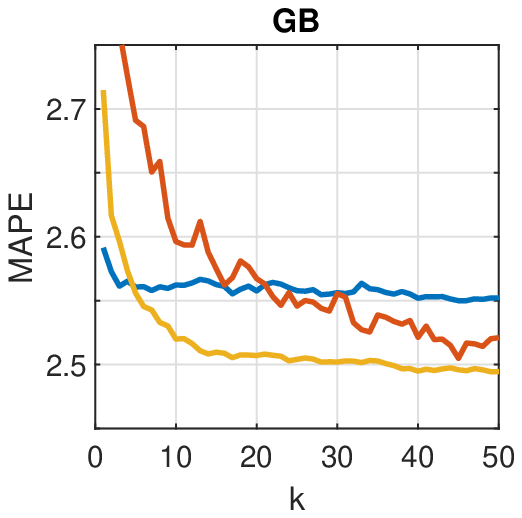}
    \includegraphics[width=0.24\textwidth]{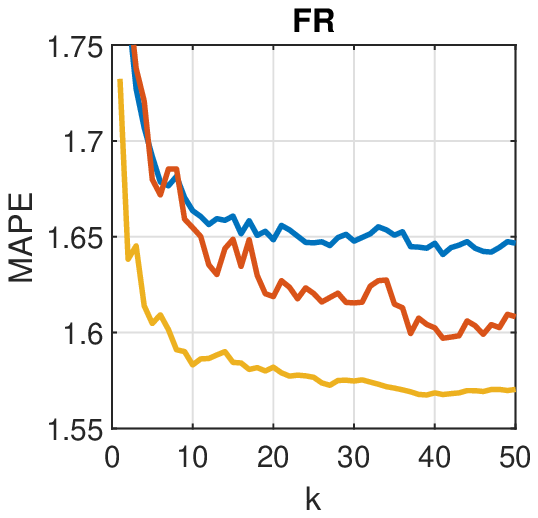}
    \includegraphics[width=0.24\textwidth]{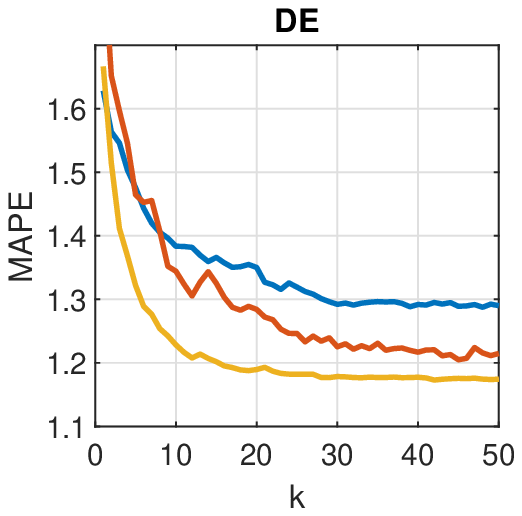}
    \caption{Ensemble error at successive stages.} 
    \label{figk}
\end{figure}

From Fig. \ref{figSens}, we can see that the accuracy of EnsR and EnsCT deteriorates quickly with the number of hidden nodes and interval $U$ size, when these hyperparameters exceed their optimal values. This deterioration is related to the gradual loss of generalization for higher values of $m$ and $\alpha_{max}$. Deterioration for EnsOR is much slower, which means that this approach is more resistant to overtraining.

Table \ref{tab1} shows optimal hyperparameters, test and training errors and compares the sensitivities of ensembling variants. The lowest values are in bold. Table  \ref{tab1} clearly shows that EnsOR performs best. This method gave the lowest errors for each dataset and had the lowest sensitivity to both hyperparameters. By contrast, EnsR performed worst in terms of error and sensitivity. Note that the optimal hyperparameter values are smaller for EnsR and EnsCT than for EnsOR. This means that the base models in these boosting variants need to be less flexible (weaker) to prevent overfitting. Nevertheless, EnsR and EnsCT do not perform as well as EnsOR. 

\begin{table}[htbp]
  \begin{center}
\setlength{\tabcolsep}{4pt}
  \caption{Optimal RandNN hyperparameters, errors and ensemble sensitivity.} \label{tab1}
    \begin{tabular}{llcccccc}
    \toprule
      Data & Ensemble & $m^*$ & $\alpha_{max}^*$ & $MAPE_{tst}$ & $MAPE_{trn}$ & $S_m$ & $S_{\alpha_{max}}$ \\
    \midrule
    \textbf{PL} 
     & EnsR  & 20    & 40               & 1.18         & 0.76         & 0.641 & 0.436              \\
     & EnsCT & 20    & 60               & 1.15         & \textbf{0.72}         & 0.092 & 0.174              \\
     & EnsOR & 40    & 70               &\textbf{ 1.14}         &\textbf{ 0.72}         & \textbf{0.064} & \textbf{0.110}              \\
    \midrule
    \textbf{GB} 
     & EnsR  & 20    & 10               & 2.55         & 1.55         & 0.509 & 1.531              \\
     & EnsCT & 10    & 50               & 2.52         & \textbf{1.51}         & \textbf{0.104} & 0.628              \\
     & EnsOR & 20    & 50               & \textbf{2.49}         & \textbf{1.51}         & \textbf{0.104} & \textbf{0.163}              \\
   \midrule
    \textbf{FR} 
     & EnsR  & 10    & 10               & 1.65         & 1.27         & 0.466 & 0.853\\
     & EnsCT & 10    & 20               & 1.61         & 1.24         & 0.088 & 0.436\\
     & EnsOR & 20    & 40               & \textbf{1.57}         &\textbf{ 1.20}         & \textbf{0.058} & \textbf{0.069}\\
    \midrule
    \textbf{DE} 
     & EnsR  & 10    & 40               & 1.29         & 1.37         & 0.774 & 0.572              \\
     & EnsCT & 10    & 70               & 1.21         & 1.26         & 0.138 & 0.141              \\
     & EnsOR & 40    & 80               & \textbf{1.17}         & \textbf{1.05}         & \textbf{0.036} & \textbf{0.063} \\
    \bottomrule
    \end{tabular}%
  \label{tab1}%
  \end{center}
\end{table}%

\begin{figure}[h]
	\centering
	\includegraphics[width=0.5\textwidth]{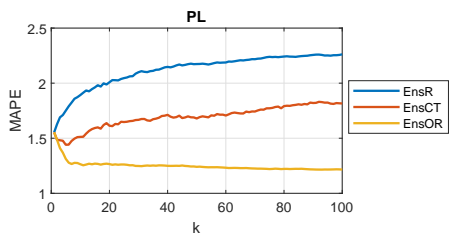}
    \caption{Ensemble error at successive stages for too flexible base models.} 
    \label{figk2}
\end{figure}

Fig. \ref{figk} demonstrates test MAPE in successive steps of ensembling. We can observe from this figure that for EnsOR the error converges faster with $k$ than for other ensemble variants. The convergence curve is also smoother for EnsOR than for its competitors. For them, in many cases, adding a new member to the ensemble 
causes a temporary increase in error. Moreover, it was observed for EnsR that if the hyperparameters are too high (greater than optimal), which means a flexible base model, the error starts to successively increase with subsequent ensembling stages. This phenomenon is to a lesser degree observed for EnsCT, but not observed for EnsOR. Fig. \ref{figk2} demonstrates this problem for PL data and $m=100$, $\alpha_{max}=70$. 

To improve further ensemble learning based on opposed response, we 
introduce weights for training patterns which express their similarity to the query pattern. The more similar training pattern $\mathbf{x}_i$ to query pattern $\mathbf{x}$, the higher the weight. The similarity measure is a scalar product, $\mathbf{x}^T\mathbf{x}_i$. The opposed response takes the form:

\begin{equation}
\hat{\mathbf{y}}'_{k,i}=\mathbf{y}_i+w_i\mathbf{r}_{k,i},   i = 1, ..., N
\label{v1}
\end{equation}
where $\mathbf{r}_{k,i}=\mathbf{y}_i-F_k(\mathbf{x}_i)$ is a residual vector for the $i$-th training pattern and $w_i \in [0, 1]$ is the weight of this pattern.

In the base variant of EnsOR, the weights for all patterns are equal to one. In typical ensemble learning, the weights are zero (the base models at each stage of ensembling learn on the original training set $\Phi$; such an approach we considered in \cite{Dud21}). By introducing weights, we try to balance these two approaches.

As a weighting function, we consider four variants. In the simplest one, EnsOR1, we assume that the weighting function $g$ is just the scalar product:

\begin{equation}
g(\mathbf{x},\mathbf{x}_i) =\mathbf{x}^T\mathbf{x}_i
\label{sp}
\end{equation}

To avoid negative values of \eqref{sp} we can also use $g(\mathbf{x},\mathbf{x}_i) =\frac{1}{2}(\mathbf{x}^T\mathbf{x}_i+1)$ or replace negative values with zeros.

In the second variant, EnsOR2, we sort the training patterns according to similarity to the query pattern, from the most to the least similar. Let $r=1, ..., N$ be the rank of the training patterns in the similarity ranking. The weighting function expresses the linear dependence of the weight on the rank:

\begin{equation}
g(\mathbf{x},\mathbf{x}_i) =1+\frac{1-r_i}{N}
\label{rr}
\end{equation}

In the third variant, EnsOR3, the weighting function expresses the non-linear dependence of the weight on the rank: 

\begin{equation}
g(\mathbf{x},\mathbf{x}_i) =\left(1+\frac{1-r_i}{N}\right)^d
\label{rr}
\end{equation}
where $d>1$ (we assume $d=4$).

In the fourth variant, EnsOR4, the most similar training patterns to the query pattern have unity weights, while the others have zero weights:

\begin{equation}
g(\mathbf{x},\mathbf{x}_i) =
\begin{cases}
    1       & \quad \text{if } \mathbf{x}_i \in \Xi_\kappa(\mathbf{x})\\
    0  & \quad \text{if } \mathbf{x}_i \notin \Xi_\kappa(\mathbf{x})
  \end{cases}
\label{nn}
\end{equation}
where $\Xi_\kappa(\mathbf{x})$ denotes a set of $\kappa$ nearest neighbors of query pattern $\mathbf{x}$ in $\Phi$.

\begin{figure}[h]
	\centering
	\includegraphics[width=0.5\textwidth]{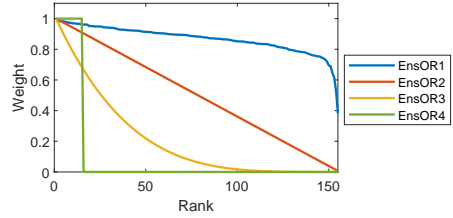}
    \caption{Examples of weights assigned to training pattern in different weighing variants.} 
    \label{figw}
\end{figure}

An example of weights assigned to the training patterns by the above weighting functions are shown in Fig. \ref{figw}.

\begin{figure}[h]
	\centering
	\includegraphics[width=0.24\textwidth]{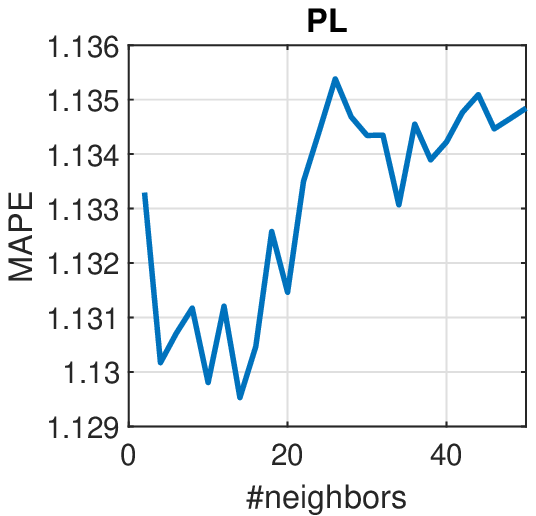}
    \includegraphics[width=0.24\textwidth]{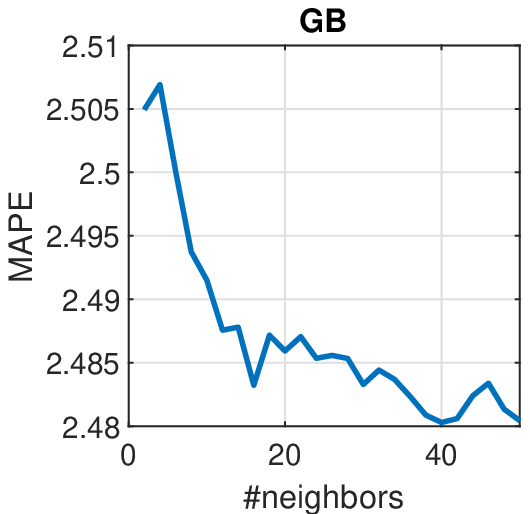}
    \includegraphics[width=0.24\textwidth]{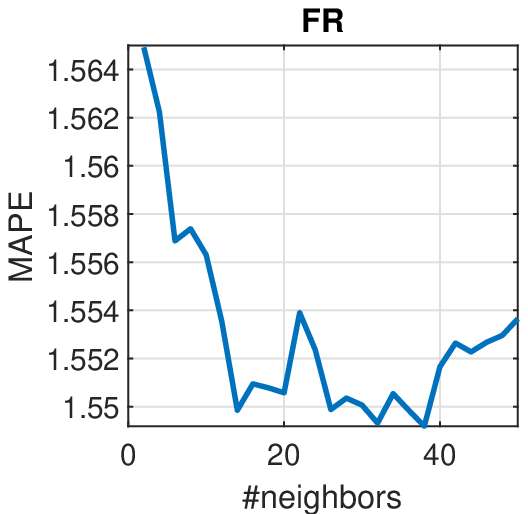}
    \includegraphics[width=0.24\textwidth]{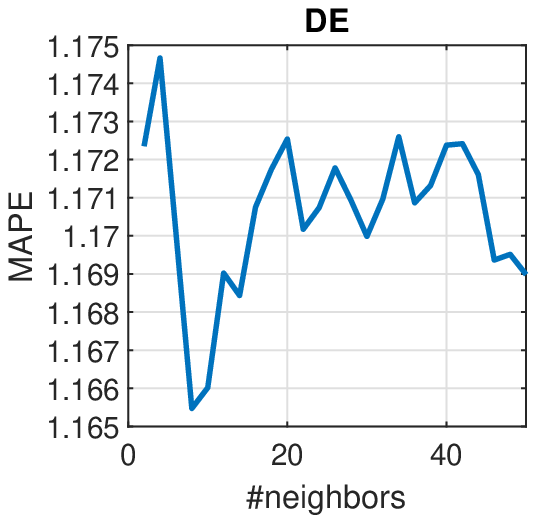}
    \caption{Ensemble error for EnsOR4 depending on $\kappa$.} 
    \label{fignn}
\end{figure}

Fig. \ref{fignn} allows us to evaluate the impact of the number of nearest neighbours $\kappa$ in EnsOR4 on the test MAPE. The optimal value of $\kappa$ depends on the data set: for PL $\kappa=14$, for GB $\kappa=40$, for FR $\kappa=38$, and for DE $\kappa=8$. 

\begin{table}[htbp]
  \begin{center}
\setlength{\tabcolsep}{4pt}
  \caption{Results for EnsOR with weighted patterns.} \label{tab1}
    \begin{tabular}{lcccccc}
    \toprule
      Data & EsnOR1 & EnsOR2 & EnsOR3 & EnsOR4 & RandNN \cite{Dud21} & Ens1 \cite{Dud21}  \\
    \midrule
    \textbf{PL} 
     & 1.1419 & 1.1381 & 1.1329 & \textbf{1.1295} & 1.3206 & 1.1417  \\
    \textbf{GB} 
     & 2.4935 & 2.4804 & 2.4822 & \textbf{2.4803} & 2.6126 & 2.5148  \\
    \textbf{FR} 
     & 1.5668 & 1.5570 & 1.5524 & \textbf{1.5492} & 1.6711 & 1.5690  \\ 
    \textbf{DE} 
     & 1.1743 & 1.1711 & 1.1683 &\textbf{1.1655} & 1.3809 & 1.1811  \\
    \bottomrule
    \end{tabular}%
  \label{tab2}%
  \end{center}
\end{table}%

Table \ref{tab2} compares the proposed methods of pattern weighting. In this table results for individual RandNN and Ens1 are also shown. Ens1 is a classical (not boosted) ensemble of RandNN. Ens1 constructs the final forecast as an average of the responses of RandNN, which learn simultaneously on the same training set $\Phi$ \cite{Dud21}. As can be seen from this table, EnsOR4 outperformed its competitors as well as classical ensembling Ens1. The much higher error for individual RandNN justifies fully ensembling. It is worth mentioning that comparing Ens1 with other forecasting models, including statistical models (ARIMA, exponential smoothing, Prophet) and machine learning models (MLP, SVM, ANFIS, LSTM, GRNN, nonparametric models), reported in \cite{Dud21} clearly shows that Ens1 outperforms all its competitors in terms of accuracy.

\section{Conclusion}

Forecasting complex time series with multiple seasonalities is a challenging problem, but one we solve using ensemble of randomized NNs. We propose three methods of boosting the RandNN ensemble. We showed that in ensemble learning based on residuals, the base models have different tasks to solve at successive stages of ensembling. Thus, the base model which is optimal at a given stage may not be optimal at other stages. To avoid having to select an optimal model at each stage of ensembling, which is unreasonable and too time-consuming, we propose to unify the tasks solved at all stages. Doing so allows us to use the same base model (the same architecture and hyperparameters), RandNN in our case, which is optimal for all stages. To unify the tasks, we propose ensemble learning based on corrected targets and ensemble learning based on opposed response. The latter proved to be more resistant to task degradation at subsequent stages.

The experimental studies performed on four forecasting problems expressing triple seasonalities confirmed that the opposed response-based approach outperforms its competitors in terms of forecasting accuracy as well as sensitivity to both the base model hyperparameters and ensemble size. Further improvement of the winning solution was achieved by weighting the training patterns according to their similarity to the query pattern.

In further research, we plan to develop the opposed response-based approach for other types of learners, e.g. decision trees, and other types of problems (regression, classification).   

%
%
%
%

\end{document}